\documentclass{article}
\let\<\textless
\let\>\textgreater
\PassOptionsToPackage{numbers, compress}{natbib}

\usepackage[final]{nips_2016}

\usepackage[utf8]{inputenc} 
\usepackage[T1]{fontenc}    
\usepackage{hyperref}       
\usepackage{url}            
\usepackage{booktabs}       
\usepackage{amsmath}
\usepackage{amsfonts}       
\usepackage{nicefrac}       
\usepackage{microtype}      
\usepackage{framed}
\usepackage[pdftex]{graphicx}
\usepackage{authblk}
\usepackage[usenames, dvipsnames]{color}

\title{A Hierarchical Latent Variable Encoder-Decoder Model for Generating Dialogues}

\renewcommand\footnotemark{}

\author[*]{Iulian V. Serban}
\author[$\ddag$]{Alessandro Sordoni \thanks{This work was carried out while A.S. was at Université de Montréal. Y.B. is a CIFAR senior Fellow.}}
\author[$\diamond$]{Ryan Lowe}
\author[$\diamond$]{Laurent Charlin}
\author[$\diamond$]{Joelle Pineau}
\author[*]{Aaron~Courville}
\author[*]{Yoshua Bengio} 
\affil[*]{\small Department of Computer Science and Operations Research, Universit{\'e} de Montr{\'e}al, Montreal, Canada \protect\\ 
          \texttt{\small \{iulian.vlad.serban,aaron.courville,yoshua.bengio\}@umontreal.ca} \protect\\
         \ } 

\affil[$\ddag$]{\small Maluuba Inc, Montreal, Canada \protect\\ 
          \texttt{\small \{alessandro.sordoni\}@maluuba.com }  \protect\\ 
            \ }

\affil[$\diamond$]{\small School of Computer Science, McGill University, Montreal, Canada \hspace*{2mm} \texttt{\small \{ryan.lowe,lcharlin,jpineau\}@cs.mcgill.ca}}

\begin{document}

\maketitle

\begin{abstract}
Sequential data often possesses a hierarchical structure with complex dependencies between subsequences, such as found between the utterances in a dialogue. 
In an effort to model this kind of generative process, we propose a neural network-based generative architecture, with latent stochastic variables that span a variable number of time steps.
We apply the proposed model to the task of dialogue response generation and compare it with recent neural network architectures.
We evaluate the model performance through automatic evaluation metrics
and by carrying out a human evaluation.
The experiments demonstrate that our model improves upon recently proposed models
and that the latent variables facilitate the generation of long outputs and maintain the context.

\end{abstract}

\section{Introduction}

Deep recurrent neural networks (RNNs) have recently demonstrated impressive results on a number of difficult machine learning problems involving the generation of sequential structured outputs~\cite{Goodfellow-et-al-2015-Book}, including language modelling~\cite{graves2012sequence,mikolov2010recurrent} machine translation~\cite{sutskever2014sequence,cho2014learning}, dialogue~\cite{sordoni2015aneural,DBLP:conf/aaai/SerbanSBCP16} and speech recognition~\cite{Goodfellow-et-al-2015-Book}.

While these advances are impressive, the underlying RNNs tend to have a fairly simple structure, in the sense that the only variability or stochasticity in the model occurs when an output is sampled.
This is often an inappropriate place to inject variability \cite{boulanger2012modeling,chung2015recurrent,bayer2014b}.
This is especially true for sequential data such as speech and natural language that possess a hierarchical generation process with complex intra-sequence dependencies. For instance, natural language dialogue involves at least two levels of structure; 
within a single utterance the structure is dominated by local statistics of the language,
while across utterances there is a distinct source of uncertainty (or variance) 
characterized by aspects such as conversation topic, speaker goals and speaker style. 

In this paper we introduce a novel hierarchical stochastic latent variable neural network architecture to explicitly model generative processes that possess multiple levels of variability.  
We evaluate the proposed model on the task of dialogue response generation and compare it with recent neural network architectures.
We evaluate the model qualitatively through manual inspection, and quantitatively using a human evaluation on Amazon Mechanical Turk and using automatic evaluation metrics.
The results demonstrate that the model improves upon recently proposed models.
In particular, the results highlight that the latent variables help to both facilitate the generation of long utterances with more information content, and to maintain the dialogue context.

\section{Technical Background}
\subsection{Recurrent Neural Network Language Model}
A recurrent neural network (RNN), with parameters $\theta$, models a variable-length sequence of tokens $(w_1, \dots, w_M)$ by decomposing the probability distribution over outputs:
\begin{align}
P_{\theta}(w_1, \dots, w_M) = \prod_{m=2}^M P_{\theta}(w_m \mid w_1, \dots, w_{m-1})P(w_1).
\end{align}
The model processes each observation recursively. 
At each time step, the model observes an element and updates its internal hidden state, $h_m  = f(h_{m-1}, w_m)$, 
where $f$ is a parametrized non-linear function, 
such as the hyperbolic tangent, the LSTM gating unit \cite{hochreiter1997long} or the GRU gating unit \cite{cho2014learning}.\footnote{We concatenate the LSTM cell and cell input hidden states into a single state $h_m$ for notational simplicity.}
The hidden state acts as a sufficient statistic, which summarizes the past sequence and parametrizes the output distribution of the model: $P_{\theta}(w_{m+1} \mid w_1, \dots, w_m) = P_{\theta}(w_{m+1} \mid h_m)$. 
We assume the outputs lie within a discrete vocabulary $V$.
Under this assumption, the RNN Language Model (RNNLM)~\cite{mikolov2010recurrent}, the simplest possible generative RNN for discrete sequences, parametrizes the output distribution using the softmax function applied to an affine transformation of the hidden state $h_m$.
The model parameters are learned by maximizing the training log-likelihood using gradient descent. 


\subsection{Hierarchical Recurrent Encoder-Decoder}

The hierarchical recurrent encoder-decoder model (HRED) \cite{sordoni2015ahier,DBLP:conf/aaai/SerbanSBCP16} is an extension of the RNNLM. 
It extends the encoder-decoder architecture \cite{cho2014learning} to the natural dialogue setting.
The HRED assumes that each output sequence can be modelled in a two-level hierarchy: sequences of sub-sequences, and sub-sequences of tokens.
For example, a dialogue may be modelled as a sequence of utterances (sub-sequences),
with each utterance modelled as a sequence of words.
Similarly, a natural-language document may be modelled as a sequence of sentences (sub-sequences),
with each sentence modelled as a sequence of words.
The HRED model consists of three RNN modules: an \textit{encoder} RNN, a \textit{context} RNN and a \textit{decoder} RNN.
Each sub-sequence of tokens is deterministically encoded into a real-valued vector by the \textit{encoder} RNN. This is given as input to the \textit{context} RNN, which updates its internal hidden state to reflect all information up to that point in time.
The \textit{context} RNN deterministically outputs a real-valued vector, which the \textit{decoder} RNN conditions on to generate the next sub-sequence of tokens.
For additional details see~\cite{sordoni2015ahier,DBLP:conf/aaai/SerbanSBCP16}.

\subsection{A Deficient Generation Process}

In the recent literature, it has been observed that the RNNLM and HRED, and similar models based on RNN architectures, have critical problems generating meaningful dialogue utterances~\cite{DBLP:conf/aaai/SerbanSBCP16,li2015diversity}.
We believe that the root cause of these problems arise from the parametrization of the output distribution in the RNNLM and HRED, which imposes a strong constraint on the generation process: the only source of variation is modelled through the conditional
output distribution.
This is detrimental from two perspectives: from a probabilistic perspective, with stochastic variations injected only at the low level, the model is encouraged to capture local structure in the sequence, rather than global or long-term structure.
This is because random variations injected at the lower level are strongly constrained to be in line with the immediate previous observations, but only weakly constrained to be in line with older observations or with future observations. One can think of random variations as injected via i.\@i.\@d.\@ noise variables, added to deterministic components, for example. If this noise is injected at a higher level of representation, spanning longer parts of the sequence, its effects could correspond to longer-term dependencies.
Second, from a computational learning perspective, the state $h_m$ of the RNNLM (or \textit{decoder} RNN of HRED) has to summarize all the past information up to time step $m$ in order to (a) generate a probable next token (short term goal) and simultaneously (b) to occupy a position in embedding space which sustains a realistic output trajectory, in order to generate probable future tokens (long term goal). Due to the vanishing gradient effect, shorter-term goals will have more influence:
finding a compromise between these two disparate forces
will likely lead the 
training procedure to model parameters that focus too much on predicting only the next output token.
In particular for high-entropy sequences, the models are very likely to favour short-term predictions as opposed to long-term predictions,
because it is easier to only learn $h_m$ for predicting the next token compared to sustaining a long-term trajectory, which at every time step is perturbed by a highly noisy source (the observed token).

\section{Latent Variable Hierarchical Recurrent Encoder-Decoder (VHRED)}


Motived by the previous discussion, we now introduce the latent variable hierarchical recurrent encoder-decoder (VHRED) model. This model augments the HRED model with a latent variable at the decoder, which is trained by maximizing a variational lower-bound on the log-likelihood. This allows it to model hierarchically-structured sequences in a two-step generation process---first sampling the latent variable, and then generating the output sequence---while maintaining long-term context.

Let $\mathbf{w}_1, \dots, \mathbf{w}_N$ be a sequence consisting of $N$ sub-sequences, 
where $\mathbf{w}_n = (w_{n,1}, \dots, w_{n,M_n})$ is the $n$'th sub-sequence 
and $w_{n,m} \in V$ is the $m$'th discrete token in that sequence. The VHRED model uses a stochastic latent variable $\mathbf{z}_n \in \mathbb{R}^{d_z}$ for each sub-sequence $n=1,\dots,N$ conditioned on all previous observed tokens. Given $\mathbf{z}_n$, the model next generates the $n$'th sub-sequence tokens $\mathbf{w}_n = (w_{n,1}, \dots, w_{n,M_n})$:
\begin{align}
P_{\theta}(\mathbf{z}_n \mid \mathbf{w}_1, \dots, \mathbf{w}_{n-1}) & = \mathcal{N}(\boldsymbol{\mu}_{\text{prior}}(\mathbf{w}_1, \dots, \mathbf{w}_{n-1}), \Sigma_{\text{prior}}(\mathbf{w}_1, \dots, \mathbf{w}_{n-1})), \\
P_{\theta}(\mathbf{w}_n \mid \mathbf{z}_n, \mathbf{w}_1, \dots, \mathbf{w}_{n-1}) & = \prod_{m=1}^{M_n} P_{\theta}(w_{n,m} \mid \mathbf{z}_n, \mathbf{w}_1, \dots, \mathbf{w}_{n-1}, w_{n,1}, \dots, w_{n,m-1}),
\end{align}
where $\mathcal{N}(\boldsymbol{\mu}, \Sigma)$ is the multivariate normal distribution with mean $\boldsymbol{\mu} \in \mathbb{R}^{d_z}$ and covariance matrix $\Sigma \in \mathbb{R}^{{d_z} \times {d_z}}$, which is constrained to be a diagonal matrix.

\begin{figure}[ht]
  \centering
  \includegraphics[width=1.0\linewidth]{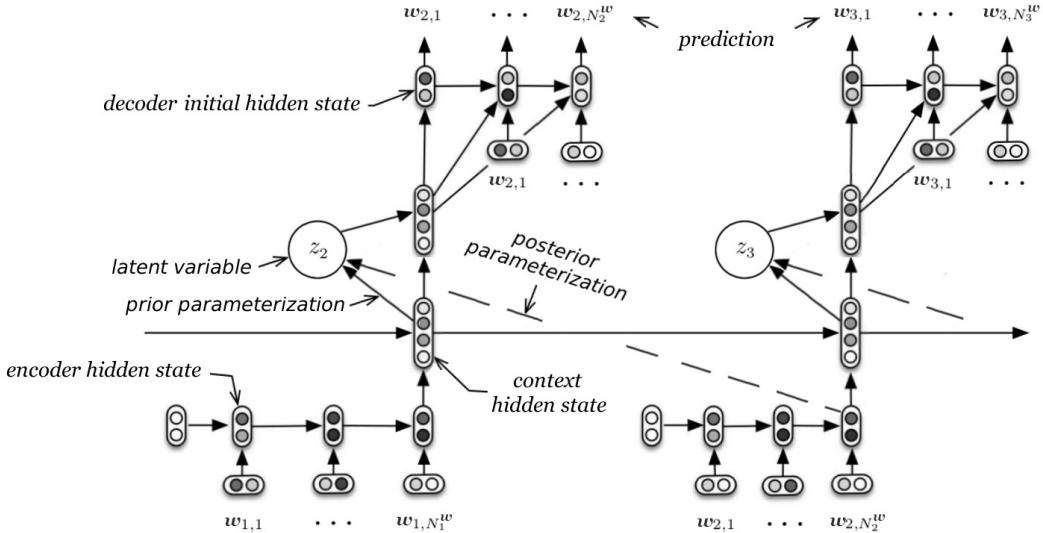}
  \caption{Computational graph for VHRED model. Rounded boxes represent (deterministic) real-valued vectors. Variables $\mathbf{z}$ represent latent stochastic variables.}
  \label{VHRED:computational_graph}
\end{figure}

The VHRED model (figure \ref{VHRED:computational_graph}) contains the same three components as the HRED model. 
The \textit{encoder} RNN deterministically encodes a single sub-sequence into a fixed-size real-valued vector.
The \textit{context} RNN deterministically takes as input the output of the \textit{encoder} RNN, and encodes all previous sub-sequences into a fixed-size real-valued vector.
This vector is fed into a two-layer feed-forward neural network with hyperbolic tangent gating function.
A matrix multiplication is applied to the output of the feed-forward network,
which defines the multivariate normal mean $\boldsymbol{\mu}_{\text{prior}}$.
Similarly, for the diagonal covariance matrix $\Sigma_{\text{prior}}$ a different matrix multiplication is applied to the net's output followed by softplus function, to ensure positiveness~\cite{chung2015recurrent}.

The model's latent variables are inferred by maximizing the variational lower-bound, which factorizes into independent terms for each sub-sequence:
\begin{align}
\log P_{\theta}(\mathbf{w}_1, \dots, \mathbf{w}_N)  \geq & \sum_{n=1}^N - \text{KL} \left [ Q_{\psi}(\mathbf{z}_n \mid \mathbf{w}_1, \dots, \mathbf{w}_n) || P_{\theta}(\mathbf{z}_n \mid \mathbf{w}_1, \dots, \mathbf{w}_{n-1}) \right ] \nonumber \\
& + \mathbb{E}_{Q_{\psi}(\mathbf{z}_n \mid \mathbf{w}_1, \dots, \mathbf{w}_n)} \left [ \log P_{\theta}(\mathbf{w}_n \mid \mathbf{z}_n, \mathbf{w}_1, \dots, \mathbf{w}_{n-1}) \right ], \label{VHRED:lower_bound}
\end{align}
where $\text{KL}[Q || P]$ is the Kullback-Leibler (KL) divergence between distributions $Q$ and $P$. The distribution $Q_{\psi}(\mathbf{z} \mid w_1, \dots, w_M)$ is the approximate posterior distribution (also known as the \textit{encoder model} or \textit{recognition model}), which aims to approximate the intractable true posterior distribution:
\begin{align}
 Q_{\psi}(\mathbf{z}_n \mid \mathbf{w}_1, \dots, \mathbf{w}_N) = Q_{\psi}(\mathbf{z}_n \mid \mathbf{w}_1, \dots, \mathbf{w}_n) & =  \mathcal{N}(\boldsymbol{\mu}_{\text{posterior}}(\mathbf{w}_1, \dots, \mathbf{w}_n), \Sigma_{\text{posterior}}(\mathbf{w}_1, \dots, \mathbf{w}_n)) \nonumber \\
 & \approx P_{\psi}(\mathbf{z}_n \mid \mathbf{w}_1, \dots, \mathbf{w}_N),
\end{align}
where $\boldsymbol{\mu}_{\text{posterior}}$ defines the approximate posterior mean and $\Sigma_{\text{posterior}}$ defines the approximate posterior covariance matrix (assumed diagonal)  as a function of the previous sub-sequences $\mathbf{w}_1, \dots, \mathbf{w}_{n-1}$ and the current sub-sequence $\mathbf{w}_{n}$. The posterior mean $\boldsymbol{\mu}_{\text{posterior}}$ and covariance $\Sigma_{\text{posterior}}$ are determined in the same way as the prior, via a matrix multiplication with the output of the feed-forward network, and with a softplus function applied for the covariance.

At test time, conditioned on the previous observed sub-sequences $(\mathbf{w}_1, \dots, \mathbf{w}_{n-1})$, a sample $\mathbf{z}_n$ is drawn from the prior $\mathcal{N}(\boldsymbol{\mu}_{\text{prior}}(\mathbf{w}_1, \dots, \mathbf{w}_{n-1}), \Sigma_{\text{prior}}(\mathbf{w}_1, \dots, \mathbf{w}_{n-1}))$ for each sub-sequence. This sample is concatenated with the output of the \textit{context} RNN and given as input to the \textit{decoder} RNN as in the HRED model, which then generates the sub-sequence token-by-token.
At training time, for $n=1,\dots,N$, a sample $\mathbf{z}_n$ is drawn from the approximate posterior $\mathcal{N}(\boldsymbol{\mu}_{\text{posterior}}(\mathbf{w}_1, \dots, \mathbf{w}_{n}), \Sigma_{\text{posterior}}(\mathbf{w}_1, \dots, \mathbf{w}_{n}))$  and used to estimate the gradient of the variational lower-bound given by Eq.~\eqref{VHRED:lower_bound}.
The approximate posterior is parametrized by its own one-layer feed-forward neural network, which takes as input the output of the \textit{context} RNN at the current time step, as well as the output of the \textit{encoder} RNN for the next sub-sequence.

The VHRED model greatly helps to reduce the problems with the generation process used by the RNNLM and HRED model outlined above.
The variation of the output sequence is now modelled in two ways:
at the sequence-level with the conditional prior distribution over $\mathbf{z}$,
and at the sub-sequence-level (token-level) with the conditional distribution over tokens $w_1,\dots, w_M$.
The variable $\mathbf{z}$ helps model long-term output trajectories,
by representing high-level information about the sequence,
which in turn allows the variable $h_m$ to primarily focus on summarizing the information up to token $M$.
Intuitively, the randomness injected by the variable $\mathbf{z}$ corresponds to higher-level decisions, like topic or sentiment of the sentence.


\section{Experimental Evaluation}
We consider the problem of conditional natural language response generation for dialogue.
This is an interesting problem with applications in areas such as customer service, technical support, language learning and entertainment \cite{young2013pomdp}.  
It is also a task domain that requires learning to generate sequences with complex structures while taking into account long-term context~\cite{lowe2015ubuntu,sordoni2015aneural}.



We consider two tasks. For each task, the model is given a dialogue context, consisting of one or more utterances, and the goal of the model is to generate an appropriate next response to the dialogue.
We first perform experiments on a \textbf{Twitter Dialogue Corpus}~\cite{ritter2011data}.
The task is to generate utterances to append to existing Twitter conversations.
The dataset is extracted using a procedure similar to Ritter et al.~\cite{ritter2011data}, and is split into training, validation and test sets,
containing respectively $749,060$, $93,633$ and $10,000$ dialogues.
Each dialogue contains $6.27$ utterances and $94.16$ tokens on average.\footnote{Due to Twitter's terms of service we are not allowed to redistribute Twitter content. Therefore, only the tweet IDs can be made public. These are available at: \url{www.iulianserban.com/Files/TweetIDs.zip}.}
The dialogues are fairly long compared to recent large-scale language modelling corpora, such as the 1 Billion Word Language Model Benchmark~\cite{DBLP:conf/interspeech/ChelbaMSGBKR14}, which focus on modelling single sentences.
We also experiment on the \textbf{Ubuntu Dialogue Corpus}~\cite{lowe2015ubuntu}, 
which contains about $500,000$ dialogues extracted from the \textit{\#Ubuntu} Internet Relayed Chat channel.
Users enter the chat channel with a Ubuntu-related technical problem,
and other users try to help them. For further details see Appendix \ref{appendix:dataset_details}.\footnote{The pre-processed Ubuntu Dialogue Corpus used is available at \url{www.iulianserban.com/Files/UbuntuDialogueCorpus.zip}.}
We chose these corpora because they are large, and have different purposes---Ubuntu dialogues are typically goal driven, where as Twitter dialogues typically contain social interaction ("chit-chat"). 

\subsection{Training and Evaluation Procedures}

We optimize all models using 
Adam \cite{kingma2014adampublished}.
We choose our hyperparameters and early stop with patience using the variational lower-bound ~\cite{Goodfellow-et-al-2015-Book}.
At test time, we use beam search with $5$ beams for outputting responses with the RNN decoders~\cite{graves2012sequence}.
For the VHRED models, we sample the latent variable $\mathbf{z}_n$,
and condition on it when executing beam search with the RNN decoder.
For Ubuntu we use word embedding dimensionality of size $300$,
and for Twitter we use word embedding dimensionality of size $400$.
All models were trained with a learning rate of $0.0001$ or $0.0002$ and with mini-batches containing $40$ or $80$ training examples.
We use a variant of truncated back-propagation and we apply gradient clipping. 
Further details are given in Appendix \ref{appendix:training_details}.


{\bf Baselines    }
On both Twitter and Ubuntu we compare to an LSTM model of $2000$ hidden units.
On Ubuntu, the HRED model has $500$, $1000$ and $500$ hidden units for the \textit{encoder}, \textit{context} and \textit{decoder} RNNs respectively. 
The \textit{encoder} RNN is a standard GRU RNN.
On Twitter, the HRED model \textit{encoder} RNN is a bidirectional GRU RNN encoder, where the forward and backward RNNs each have $1000$ hidden units,
and \textit{context} RNN and \textit{decoder} RNN have each $1000$ hidden units.
For reference, we also include a non-neural network baseline, specifically the TF-IDF retrieval-based model proposed in \cite{lowe2015ubuntu}.

{\bf VHRED    }
The \textit{encoder} and \textit{context} RNNs for the VHRED model are parametrized in the same way as the corresponding HRED models.
The only difference in the parametrization of the \textit{decoder} RNN is that the 
\textit{context} RNN output vector is now concatenated with the generated stochastic latent variable.
Furthermore, we initialize the feed-forward networks of the prior and posterior distributions with values drawn from a zero-mean normal distribution with variance $0.01$ and with biases equal to zero.
We also multiply the diagonal covariance matrices of the prior and posterior distributions with $0.1$ to make training more stable, because a high variance makes the gradients w.r.t. the reconstruction cost unreliable, which is fatal at the beginning of the training process.

The VHRED's \textit{encoder} and \textit{context} RNNs are initialized to the parameters of the corresponding converged HRED models.
We also use two heuristics proposed by Bowman et al.~\cite{bowman2015generating}: we drop words in the decoder with a fixed drop rate of $25\%$ and multiply the KL terms in eq.~\eqref{VHRED:lower_bound} by a scalar, which starts at zero and linearly increases to $1$ over the first $60,000$ and $75,000$ training batches on Twitter and Ubuntu respectively.
Applying these heuristics helped substantially to stabilize the training process and make the model use the stochastic latent variables.
We experimented with the batch normalization training procedure for the feed-forward neural networks.
but found that this made training very unstable without any substantial gains in performance w.r.t.\@ the variational bound.


{\bf Evaluation    }
Accurate evaluation of dialogue system responses is a difficult problem~\cite{GalleyBSJAQMGD15,pietquin2013survey}. 
Inspired by metrics for machine translation and information retrieval, researchers have begun adopting word-overlap metrics, however Liu et al.~\cite{liu2016not} show that such metrics have little correlation with human evaluations of response quality. 
We therefore carry out a human evaluation to compare responses from the different models.
We also compute several statistics and automatic metrics on model responses to characterize differences between the model-generated responses.

We carry out the human study for the Twitter Dialogue Corpus on Amazon Mechanical Turk (AMT). We do not conduct AMT experiments on Ubuntu as evaluating these responses usually requires technical expertise, which is not prevalent among AMT users.
%
We set up the evaluation study as a series of pairwise comparison experiments.\footnote{Source code for the AMT experiments will be released upon publication.}
We show human evaluators a dialogue context along with two potential responses, one generated from each model (conditioned on dialogue context). We ask 
participants to choose the response most appropriate to the dialogue context.
If the evaluators are indifferent to either of the two responses, or if they cannot understand the dialogue context, they can choose neither response.
For each pair of models we conduct two experiments: one where the example contexts contain at least $80$ unique tokens (\emph{long context}),
and one where they contain at least $20$ (not necessarily unique) tokens (\emph{short context}).
This helps compare how well each model can integrate the dialogue context into its response, since it has previously been hypothesized that for long contexts hierarchical RNNs models fare better~\cite{DBLP:conf/aaai/SerbanSBCP16,sordoni2015ahier}.
Screenshots and further details of the experiments are  in Appendix \ref{appendix:amt}.

\subsection{Results of Human Evaluation}

\begin{table}[t]
  \caption{Wins, losses and ties (in \%) of the VHRED model against the baselines based on the human study on Twitter (mean preferences $\pm \ 90\%$ confidence intervals) }
  \label{table:human-study-twitter}
  \def\arraystretch{1}
  \small
  \centering
  \begin{tabular}{lccccccc}
    \toprule
         & \multicolumn{3}{c}{\textbf{Short Contexts}} & & \multicolumn{3}{c}{\textbf{Long Contexts}} \\
         \midrule
    \textbf{Opponent} & \textbf{Wins} & \textbf{Losses} &  \textbf{Ties} & &\textbf{Wins} & \textbf{Losses} & \textbf{Ties} \\ \midrule
    VHRED vs LSTM & $32.3 \pm 2.4$ & $\mathbf{42.5 \pm 2.6}$ & $25.2 \pm 2.3$ & & $\mathbf{41.9 \pm 2.2}$ & $36.8 \pm 2.2$ & $21.3 \pm 1.9$ \\
    VHRED vs HRED & $\mathbf{42.0 \pm 2.8}$ & $31.9 \pm 2.6$ & $26.2 \pm 2.5$ & & $\mathbf{41.5 \pm 2.8}$ & $29.4 \pm 2.6$ & $29.1 \pm 2.6$ \\ \midrule
    VHRED vs TF-IDF & $\mathbf{51.6 \pm 3.3}$ & $17.9 \pm 2.5$ & $30.4 \pm 3.0$ & & $\mathbf{47.9 \pm 3.4}$ & $11.7 \pm 2.2$ & $40.3 \pm 3.4$ \\ \bottomrule
  \end{tabular}
 \end{table}

\begin{table}[t]
  \caption{Twitter examples for the neural network models. The $\rightarrow$ token indicates a change of turn.} 
  \label{table:twitter-examples}
  \scriptsize
  \centering
  \begin{tabular}{p{60mm}|p{75mm}}
    \textbf{Context} & \textbf{Response} \\ \hline
    { \color{Black} jamaica blood!! <url> } $\rightarrow$ { \color{Blue} lol. yes [name]!. [name]! } $\rightarrow$ { \color{Black} what's poppin in toronto homie } & \textbf{VHRED:} nothing much. just chillin. what's good with you? \newline \textbf{LSTM:} \ something!!!!!!! \newline \textbf{HRED:} something!!!!!!   \\ \hline
    { \color{Black} whhhhhhyyyy is it suddenly cold in tallahassee today?? this is florida! how can i dream of the beach when my fingers are icey cold?? } $\rightarrow$ { \color{Blue} sittin on deck in sun looking at lake travis in austin! enjoying day b4 birthday} & \textbf{VHRED:} the sun is shining!!! \newline \textbf{LSTM:} i know!!!!!!!!! \newline \textbf{HRED:} something!!!!!!!   \\ \hline
    { \color{Black} reaches for [name]! } $\rightarrow$ { \color{Blue} - approaches him warily - hey there buddy. - swings him into the air - how you doin? } $\rightarrow$ { \color{Black} tilts his head and your wariness and reaches for your face [name]. } & \textbf{VHRED:} - looks up at you and smiles - you're so cute. - kisses your cheek - \newline \textbf{LSTM:} \ - smiles - i'm fine thank you. how are you? \newline \textbf{HRED:} - grins, looking at you as i look at you -   \\ \hline
    { \color{Black} i think i want an ipad2. but i already have an ibook. want vs. need.} $\rightarrow$ { \color{Blue} if you have an ibook. then you deserve an ipad! } & \textbf{VHRED:} that's what i'm thinking. i'm going to buy a new one.\newline \textbf{LSTM:} i want one!!!! \newline \textbf{HRED:} i don't know what to do with it. i don't know what to do. \\ \hline
    { \color{Black}  ah echt? dan anrtwoord ik volgende keer via twitter als je me smst. } $\rightarrow$  { \color{Blue}  haha doe maar p. goed bezig hoor myrthe d } $\rightarrow$ { \color{Black}  goed bezig? die snap ik even niet p } & \textbf{VHRED:}  haha, ik heb er geen zin in de something, maar ik heb er geen something meer something \newline \textbf{LSTM:}  haha something something something  \newline \textbf{HRED:} something something something something something something something something \\ \hline
    { \color{Black}  i love katie waissel more than like, anything in life. } $\rightarrow$  { \color{Blue}  i wanna c the show again! i need <number> meet her again! n i want a b-day tweet nxt week doubt that will happen x } & \textbf{VHRED:} me too. i'm going to have to wait for her to come back. i'm so excited. \newline \textbf{LSTM:} i'm sure she will. i'm sure she'll be fine. \newline \textbf{HRED:} i'm sure she'll be fine, i'm sure she'll be fine.\\ \hline
    { \color{Black}  my math teacher over here talkn bout nickels nd i'm tryna make big money i need to be in calculus } $\rightarrow$ { \color{Blue} lol } & \textbf{VHRED:} i need to go to the gym and get my hair done lol \newline \textbf{LSTM:} lol \newline \textbf{HRED:} how u been \\
    \bottomrule
  \end{tabular}
\end{table}

The results (table~\ref{table:human-study-twitter}) show that VHRED is clearly preferred in the majority of the experiments.
In particular, VHRED is strongly preferred over the HRED and TF-IDF baseline models for both short and long context settings. 
VHRED is also preferred over the LSTM baseline model for long contexts; however, the LSTM is preferred over VHRED for short contexts.
We believe this is because the LSTM baseline tends to output much more \textit{generic} responses (see table \ref{table:response-information-content}); since it doesn't model the hierarchical input structure, the LSTM model has a shorter memory span, and thus must output a response based primarily on the end of the last utterance. 
Such `safe' responses are reasonable for a wider range of contexts, meaning that human evaluators are more likely to rate them as appropriate. However, we argue that a model that only outputs generic responses is undesirable for dialogue, as this leads to uninteresting and less engaging conversations. 
Conversely, the VHRED model is explicitly designed for long contexts, and to output a diverse set of responses due to the sampling of the latent variable.
Thus, the VHRED model generates longer sentences with more semantic content than the LSTM model (see tables~\ref{table:one-utterance-paraphrase-similarity}-\ref{table:response-information-content}). This can be `riskier' as longer utterances are more likely to contain small mistakes, which can lead to lower human preference for a single utterance. However, we believe that response diversity is crucial to maintaining interesting conversations --- in the dialogue literature, generic responses are used primarily as `back-off' strategies in case the agent has no interesting response that is relevant to the context~\cite{shaikh2010vca}. 


The above hypotheses are confirmed upon qualitative assessment of the generated responses (table \ref{table:twitter-examples}). 
VHRED generates longer and more meaningful responses compared to the LSTM model, which generates mostly generic responses. 
Additionally, we observed that the VHRED model has learned to better model smilies, slang (see first example in table \ref{table:twitter-examples}) and can even continue conversations in different languages (see fifth example).\footnote{There is a notable amount of Spanish and Dutch conversations in the corpus.}
Such aspects are not measured by the human study.
Further, VHRED appears to be better at generating \textit{stories} or \textit{imaginative actions} compared to the generative baseline models (see third example).
The last example in table \ref{table:twitter-examples} is a case where the VHRED generated response is more interesting, yet may be less preferred by humans as it is slightly incompatible with the context, compared to the generic LSTM response.
In the next section, we back these examples quantitatively, showing that the VHRED model learns to generate \textit{longer} responses with \textit{more information content} that \textit{share semantic similarity} to the context and ground-truth response.

\subsection{Results of Metric-based Evaluation}

\begin{table}[t]
  \def\arraystretch{1}
  \caption{Evaluation on 1-turn and 3-turns dialogue generation using the proposed embedding metrics.}
  \label{table:one-utterance-paraphrase-similarity}
  \small
  \centering
  \begin{tabular}{lccccccc}
    \toprule
     & \multicolumn{3}{c}{\textbf{Twitter}} & & \multicolumn{3}{c}{\textbf{Ubuntu}} \\
     \midrule
    \textbf{Model} & \textbf{Average} & \textbf{Greedy} &  \textbf{Extrema} & & \textbf{Average} & \textbf{Greedy} &  \textbf{Extrema} \\
    \midrule
     & \multicolumn{7}{c}{\textbf{1-turn}} \\
    \midrule
    LSTM & $0.512$ & $0.389$ & $0.366$ & & $0.23$ & $0.169$ & $0.157$ \\
    HRED & $0.501$ & $0.378$ & $0.355$ & & $\mathbf{0.577}$ & $\mathbf{0.417}$ & $\mathbf{0.391}$ \\
    VHRED & $\mathbf{0.533}$ & $\mathbf{0.396}$ & $\mathbf{0.38}$ & & $0.542$ & $0.384$ & $0.363$ \\
    \midrule
    & \multicolumn{7}{c}{\textbf{3-turns}} \\
    \midrule
    LSTM & $0.657$ & $0.561$ & $0.374$ &  &$0.638$ & $0.456$ & $0.378$ \\
    HRED & $0.646$ & $0.552$ & $0.364$ &  &$0.742$ & $0.524$ & $0.432$ \\ 
    VHRED & $\mathbf{0.689}$ & $\mathbf{0.583}$ & $\mathbf{0.391}$ & & $\mathbf{0.777}$ & $\mathbf{0.536}$ & $\mathbf{0.448}$ \\ \bottomrule
  \end{tabular}
\end{table}

To show the VHRED responses are more on-topic and share semantic similarity to the ground-truth response, we consider three textual similarity metrics based on word embeddings.
The \textit{Embedding Average} (Average) metric projects the model response and ground truth response into two separate real-valued vectors by taking the mean over the word embeddings in each response, and then computes the cosine similarity between them~\cite{mitchell2008vector}. 
This metric is widely used for measuring textual similarity.
The \textit{Embedding Extrema} (Extrema) metric similarly embeds the responses by taking the extremum (maximum of the absolute value) of each dimension, 
and afterwards computes the cosine similarity between them.
The \textit{Embedding Greedy} (Greedy) metric is more fine-grained; 
it uses cosine similarity between word embeddings to find the closest word
in the human-generated response for each word in the model response.
Given the (non-exclusive) alignment between words in the two responses,
the mean over the cosine similarities is computed for each pair of questions~\cite{Rus:2012:CGO:2390384.2390403}. 
Since this metric takes into account the alignment between words, it should be more accurate for long responses.
While these metrics do not strongly correlate with human judgements of generated responses, we interpret them as 
measuring \textit{topic similarity}: if the model generated response has similar semantic content to the ground truth human response, then the metrics will yield a high score.
To ease reproducibility, we use the publicly available Word2Vec word embeddings trained on the Google News Corpus.\footnote{\url{https://code.google.com/archive/p/word2vec/}} 

We compute these metrics in two settings: one where the models generate a single response (1-turn), and one where they generate the next three consecutive utterances (3-turns) (table \ref{table:one-utterance-paraphrase-similarity}). 
Overall, VHRED seems to better capture the ground truth response topic than either the LSTM or HRED models.
The fact that VHRED does better in particular in the setting where the model generates three consecutive utterances strongly suggests that hidden states in both the \textit{decoder} and \textit{context} RNNs of the VHRED models are better able to  follow trajectories which remain on-topic w.r.t\@ the dialogue context.
This supports our computational hypothesis that the stochastic latent variable helps modulate the training procedure to achieve a better trade-off between short-term and long-term generation.
We also observed the same trend when computing the similarity metrics between the model generated responses and the corresponding context, which further reinforces this hypothesis.

\begin{table}[t]
  \caption{Response information content on 1-turn generation as measured by average utterance length $|U|$, word entropy $H_w = -\sum_{w \in U} p(w) \log p(w)$ and utterance entropy $H_U$ with respect to the maximum-likelihood unigram distribution of the training corpus $p$.}
  \label{table:response-information-content}
  \small
  \centering
  \begin{tabular}{lcccccccc}
    \toprule
    & \phantom{a} & \multicolumn{3}{c}{\textbf{Twitter}} & \phantom{ab} & \multicolumn{3}{c}{\textbf{Ubuntu}} \\ 
    \midrule
    \textbf{Model} & & \textbf{$|U|$} & \textbf{$H_w$} &  \textbf{$H_U$} & & \textbf{$|U|$} & \textbf{$H_w$} &  \textbf{$H_U$} \\
    \midrule
    LSTM & & $11.21$ & $6.75$ &  $75.61$ & & $4.27$ & $6.50$ & $27.77$ \\
    HRED & & $11.64$ & $6.73$ &  $78.35$ & & $\mathbf{11.05}$ & $7.53$ & $\mathbf{83.16}$ \\ 
    VHRED & & $\mathbf{12.29}$ & $\mathbf{6.88}$ & $\mathbf{84.56}$ & & $9.22$ & $\mathbf{7.70}$ & $71.00$ \\  \midrule
    Human &  & $20.57$ & $8.10$ &$166.57$ & & $18.30$ & $8.90$ & $162.88$ \\ \bottomrule
  \end{tabular}
 \end{table}

To show that the VHRED responses contain more information content than other model responses, 
we compute the average response length and average entropy (in bits) w.r.t.\@ the maximum likelihood unigram model over the generated responses (table \ref{table:response-information-content}).
The unigram entropy is computed on the preprocessed tokenized datasets.
VHRED produces responses with higher entropy per word on both Ubuntu and Twitter compared to the HRED and LSTM models.
VHRED also produces longer responses overall on Twitter,
which translates into responses containing on average $~6$ bits of information more than the HRED model.
Since the actual dialogue responses contain even more information per word than any of the generative models, it reasonable to assume that a higher entropy is desirable.
Thus, VHRED compares favourably to recently proposed models in the literature, which often output extremely low-entropy (generic) responses
such as \textit{OK} and \textit{I don't know}~\cite{DBLP:conf/aaai/SerbanSBCP16,li2015diversity}.
Finally, the fact that VHRED produces responses with higher entropy suggests that its responses are on average more diverse than the responses produced by the HRED and LSTM models.
This implies that the trajectories of the hidden states of the VHRED model traverse a larger area of the space compared to the hidden states of the HRED and LSTM baselines,
which further supports our hypothesis that the stochastic latent variable helps the VHRED model achieve a better trade-off between short-term and long-term generation.


\section{Related Work}


The use of a stochastic latent variable learned by maximizing a variational lower bound is inspired by the variational autoencoder (VAE)~\cite{kingma2013auto,rezende2014stochastic}. Such models have been used predominantly for generating images in the continuous domain~\cite{gregor2015draw}. 
However, there has also been recent work applying these architectures for generating sequences, such as the Variational Recurrent Neural Networks (VRNN)~\cite{chung2015recurrent},
which was applied for speech and handwriting synthesis,
and Stochastic Recurrent Networks (STORN)~\cite{bayer2014b},
which was applied for music generation and motion capture modeling.
Both the VRNN and STORN incorporate stochastic latent variables into RNN architectures, but unlike the VHRED they sample a separate latent variable at each time step of the decoder.
This does not exploit the hierarchical structure in the data, and thus does not model higher-level variability.




Similar to our work is the Variational Recurrent Autoencoder~\cite{fabius2014variational} and the Variational Autoencoder Language Model~\cite{bowman2015generating}, which apply encoder-decoder architectures to generative music modeling and language modeling respectively.
The VHRED model is different from these in the following ways.
The VHRED latent variable is conditioned on all previous sub-sequences (sentences). This enables the model to generate multiple sub-sequences (sentences),
but it also makes the latent variables co-dependent through the observed tokens.
The VHRED model builds on the hierarchical architecture of the HRED model, which makes the model applicable to generation conditioned on long contexts.
It has a direct deterministic connection between the \textit{context} and \textit{decoder} RNN, which allows the model to transfer deterministic pieces of information between its components.\footnote{Our initial experiments confirmed that the deterministic connection between the \textit{context} RNN to the \textit{decoder} RNN was indeed beneficial in terms of lowering the variational bound.}
Crucially, VHRED also demonstrates improved results beyond the autoencoder framework, where the objective is not input reconstruction but the conditional generation of the next utterance in a dialogue.

\newpage

\section{Discussion}
We have introduced a novel latent variable neural network architecture, called VHRED.
The model uses a hierarchical generation process in order to exploit the structure in sequences and is trained using a variational lower bound on the log-likelihood.
We have applied the proposed model on the difficult task of dialogue response generation, and have demonstrated that it is an improvement over previous models in several ways, including quality of responses as measured in a human study.
The empirical results highlight the advantages of the hierarchical generation process for modelling high-entropy sequences.
Finally, it is worth noting that the proposed model is very general.
It can in principle be applied to any sequential generation task that exhibits a hierarchical structure, such as document-level machine translation, web query prediction, 
multi-sentence document summarization,
multi-sentence image caption generation, and others.

\bibliographystyle{natbib}
\begingroup
    \vspace{-2.0mm}
    \footnotesize
    \setlength{\bibsep}{3pt}
    \bibliography{ref}
\endgroup

\small
\newpage
\section*{Appendix}

\subsection{Dataset Details} \label{appendix:dataset_details}

Our Twitter Dialogue Corpus was extracted in $2011$. 
We perform a minimal preprocessing on the dataset to remove irregular punctuation marks and tokenize it using the Moses tokenizer: \url{https://github.com/moses-smt/mosesdecoder/blob/master/scripts/tokenizer/tokenizer.perl}.

We use the Ubuntu Dialogue Corpus v2.0 extracted in Jamuary 2016 from: \url{http://cs.mcgill.ca/~jpineau/datasets/ubuntu-corpus-1.0/}. The preprocessed version of the dataset will be made available to the public.

\subsection{Model Details} \label{appendix:training_details}

The model implementations will be released to the public upon acceptance of the paper.

\subsubsection*{Training and Generation}
We validate each model on the entire validation set every $5000$ training batches.

As mentioned in the main text, at test time we use beam search with $5$ beams for outputting responses with the RNN decoders~\cite{graves2012sequence}.
We define the beam cost as the log-likelihood of the tokens in the beam divided by the number of tokens it contains. This is a well-known modification, which is often applied in machine translation models. 
In principle, we could sample from the RNN decoders of all the models,
but is well known that such sampling produces poor results in comparison to the beam search procedure.
It also introduces additional variance into the evaluation procedure, which will make the human study very expensive or even impossible within a limited budget.

\subsubsection*{Baseline Models}
On Ubuntu, the gating function between the \textit{context} RNN and \textit{decoder} RNN is a one-layer feed-forward neural network with hyperbolic tangent activation function.

On Twitter, the HRED \textit{decoder} RNN computes a $1000$ dimensional real-valued vector for each hidden time step, which is multiplied with the output \textit{context} RNN. The result is feed through a one-layer feed-forward neural network with hyperbolic tangent activation function, which the \textit{decoder} RNN then takes as input.
Furthermore, the \textit{encoder} RNN initial state for each utterance is initialized to the last hidden state of the \textit{encoder} RNN from the previous utterance.
We found that this worked slightly better for the VHRED model, but a more careful choice of hyperparameters is likely to make this additional step unnecessary for both the HRED and VHRED models.

\subsubsection*{Latent Variable Parametrization}
We here describe the formal definition of the latent variable prior and approximate posterior distributions.
Let $w_1,\dots,w_T$ be discrete tokens in vocabulary $V$, which correspond to one sequence (e.g.\@ one dialogue).
Let $h_{t,con} \in \mathbb{R}^{d_{h,con}}$ be the hidden state of the HRED \textit{context} encoder at time $t$.
Then the prior mean and covariance matrix are given as:
\begin{align}
\bar{h}_{t,con} &= \text{tanh}(H_{l_2,prior} \text{tanh}(H_{l_1,prior}h_{t,con} + b_{l_1,prior}) + b_{l_2,prior} ), \\
\boldsymbol{\mu}_{t,\text{prior}} &= H_{\mu,prior} \bar{h}_{t,con} + b_{\mu,prior}, \\
\Sigma_{t,\text{prior}} &= \text{diag} ( \log (1 + \exp ( H_{\Sigma,prior} \bar{h}_{t,con} + b_{\Sigma,prior}))),
\end{align}
where the parameters are $H_{l_1,prior} \in \mathbb{R}^{d_z \times d_{h,con}}$, $H_{\Sigma,prior}, H_{\mu,prior}, H_{l_2,prior} \in \mathbb{R}^{d_z \times d_z}$ and $b_{l_1,prior}, b_{l_2,prior}, b_{\mu,prior}, b_{\Sigma,prior} \in \mathbb{R}^{d_z}$, and where $\text{diag}(\mathbf{x})$ is a function mapping a vector $\mathbf{x}$ to a matrix with diagonal elements $\mathbf{x}$ and all off-diagonal elements equal to zero.
At generation time, the latent variable is sampled at the end of each utterance:
$\mathbf{z}_t \sim \mathcal{N}(\boldsymbol{\mu}_{t.\text{prior}}, \Sigma_{t,\text{prior}})$.

The equations for the approximate posterior are similar.
Let $h_{t,p} \in \mathbb{R}^{d_{h,con}+d_{h,enc}}$ be the concatenation of $h_{t,con}$
and the hidden state of the \textit{encoder} RNN at the end of the next sub-sequence, which we assume has dimensionality $d_{h,enc}$. 
The approximate posterior mean and covariance matrix are given as:
\begin{align}
\bar{h}_{t,p} &= \text{tanh}(H_{l_2,posterior} \text{tanh}(H_{l_1,posterior}h_{t,p} + b_{l_1,posterior})) + b_{l_2,posterior}) ), \\
\boldsymbol{\mu}_{t,\text{posterior}} &= H_{\mu,posterior} \bar{h}_{t,p} + b_{\mu,posterior}, \\
\Sigma_{t,\text{posterior}} &= \text{diag} ( \log (1 + \exp ( H_{\Sigma,posterior} \bar{h}_{t,p} + b_{\Sigma,posterior}) )),
\end{align}
where $H_{l_1,posterior} \in \mathbb{R}^{d_z \times (d_{h,con} + d_{h,enc})}$, $H_{\Sigma,posterior}, H_{\mu,posterior}, H_{l_2,posterior} \in \mathbb{R}^{d_z \times d_z}$ and $b_{l_1,posterior}, b_{l_2,posterior}, b_{\mu,posterior}, b_{\Sigma,posterior} \in \mathbb{R}^{d_z}$ are its parameters.
The derivative of the variational bound is computed by sampling one latent variable at the end of each utterance: $\mathbf{z}_t \sim \mathcal{N}(\boldsymbol{\mu}_{t.\text{posterior}}, \Sigma_{t,\text{posterior}})$.

\newpage

\subsection{Model Examples} \label{appendix:model_examples}

Model responses for Ubuntu are shown in Table \ref{table:ubuntu-examples}.
All model responses are available for download at \url{www.iulianserban.com/Files/UbuntuDialogueCorpus.zip} and \url{http://www.iulianserban.com/Files/TwitterDialogueCorpus.zip}.

\begin{table}[ht]
  \caption{Ubuntu model examples. The $\rightarrow$ token indicates a change of turn.}
  \label{table:ubuntu-examples}
  \scriptsize
  \centering
  \begin{tabular}{p{100mm}|p{35mm}}
    \textbf{Context} & \textbf{Response} \\ \hline
        i set up my hd such that i have to type a passphrase to access it at boot. how can i remove that passwrd, and just boot up normal. i did this at install, it works fine, just tired of having reboots where i need to be at terminal to type passwd in. help?  $\rightarrow$ backup your data, and re-install without encryption " might " be the easiest method  & \textbf{VHRED:} problem is that i am not able to boot into the live cd, i can access it from the live cd, but i dont know how to use it \newline \textbf{HRED:} how would i do that? \newline \textbf{LSTM:} how do i do that? \\ \hline
        hi, everyone. I download the googleearth. deb, and want to install it with : sudo dpkg -i something. deb, It tells me : google-earth-stable depends on lsb-core (>= 3.2 ) ; however : Package lsb-core is not installed. So i just install the lsb-core, But it comes error : <url>   what should i do?  $\rightarrow$ the error suggests running " sudo apt-get -f install "  $\rightarrow$ i tried that, it comes the same error  & \textbf{VHRED:} pastebin your /etc/apt/sources. list file \newline \textbf{HRED:} try " sudo apt-get -f install " \newline \textbf{LSTM:} something \\ \hline
        I've a 2.0 TB hard drive. It no longer boots from the hard drive, but I can see it from Gpart and it has a " boot " flag.  How do I recover it?  $\rightarrow$ the boot flag doesn't mean anything in Linux  why do you not have a backup?  you can mount the partition in liveCD and access the data easily ( assuming the partition is healthy )  $\rightarrow$ I have one hard drive for the moment ( well, I don't count the 250GB drive ) besides, I don't have two SATA cables  I will make a backup as soon as I get another nice sized hard drive and a SATA cable  And how do I find out if the partition is healthy?  $\rightarrow$ boot to ubuntu livecd and try and mount the partition  & \textbf{VHRED:} fdisk -l gives me a list of partitions \newline \textbf{HRED:} How do I do that? \newline \textbf{LSTM:} on the live cd? \\ \hline
        bottom line is usually if you got 64bit why not use it? ( unless there is something that doesn't work on 64bit, which is amitedly not much )  $\rightarrow$ well, now Clip2net is not working... just seems Ubuntu runs more seamlessly on 32-bit and after reading this article, Im wondering if 64 is a bit hyped? <url>  & \textbf{VHRED:} there is a 64 bit version of ubuntu \newline \textbf{HRED:} I have no experience with 64bit, so I don't know what you mean. \newline \textbf{LSTM:} something, something, something, something, something \\ \hline
        Its comming from my computer  15 people using same router and they have no issues, also it happened recently, before it was ok  $\rightarrow$ yes but your router may have optional settings for dns which if not set to the same dns will override your network interface... it 's an option to consider  & \textbf{VHRED:} I know, I was just trying to help you. \newline \textbf{HRED:} how do I do that? \newline \textbf{LSTM:} that is what I was looking for, thanks \\ \hline
        it looks as if the network 's DHCP server is dead : " < warn > ( eth0) : DHCPv4 request timed out "  $\rightarrow$ but it isn't.  windows works  & \textbf{VHRED:} can you ping the router? \newline \textbf{HRED:} that 's what I thought, but I don't know how to fix it \newline \textbf{LSTM:}? \\ \hline
        I've been having trouble with a new 3TB harddrive and a fresh Ubuntu install. I install ( as GPT ) and the install looks ok. But then get GRUB " error : out of disk " after reboot. I have an October 2011 PC with updated BIOS and I \_do\_ see the 3TB.  google searches haven't been to helpful so far.  has anyone seen GRUB trouble when installing on 3TB harddrives?  $\rightarrow$ have you tried installing grub again and updating it? sudo grub-install \; sudo update-grub?  $\rightarrow$ I guess I could try that. So I would have to install GRUB from an older Ubuntu install?  & \textbf{VHRED:} i have no idea, sorry \newline \textbf{HRED:} yes, you can install grub to the MBR of your hard drive \newline \textbf{LSTM:} grub is installed on the mbr of the hard drive \\ \hline
    \bottomrule
  \end{tabular}
\end{table}

\newpage

\subsection{Human Study on Amazon Mechanical Turk} \label{appendix:amt}

\subsubsection*{Setup}

We choose to use crowdsourcing platforms such as AMT rather than carrying out in-lab experiments, even though in-lab experiments usually exhibit less noise and result in higher agreement between human annotators.
We do this because AMT experiments involve a larger and more heterogeneous pool of annotators, which implies less cultural and geographic biases,
and because such experiments are easier to replicate, which we believe is important for benchmarking future research on these tasks. 

Allowing the AMT human evaluators to not assign preference for either response is important, since there are several reasons why humans may not understand the dialogue context, which include topics they are not familiar with, slang language and non-English language.
We refer to such evaluations as `indeterminable'.

The evaluation setup resembles the classical Turing Test 
where human judges have to distinguish between human-human conversations and human-computer conversations.
However, unlike the original Turing Test, we only ask human evaluators to consider the next utterance in a given conversation and we do not inform them that any responses were generated by a computer.
Apart from minimum context and response lengths we impose no restrictions on the generated responses.

\subsubsection*{Selection Process}

At the beginning of each experiment, we briefly instruct the human evaluator on the task and show them a simple example of a dialogue context and two potential responses.
To avoid presentation bias, we shuffle the order of the examples and the order of the potential responses for each example.
During each experiment, we also show four trivial `attention check' examples that any human evaluator who has understood the task should be able to answer correctly.
We discard responses from human evaluators who fail more than one of these checks.

We select the examples shown to human evaluators at random from the test set. 
We filter out all non-English conversations and conversations containing offensive content.
This is done by automatically filtering out all conversations with non-ascii characters and conversations with profanities, curse words and otherwise offensive content.
This filtering is not perfect, so we manually skim through many conversations and filter out conversations with non-English languages and offensive content.
On average, we remove about $1/80$ conversations manually.
To ensure that the evaluation process is focused on evaluating conditional dialogue response generation (as opposed to unconditional single sentence generation), we constrain the experiment by filtering out examples with fewer than $3$ turns in the context.
We also filter out examples where either of the two presented responses contain less than $5$ tokens.
We remove the special token placeholders and apply regex expressions to detokenize the text.


\subsubsection*{Execution}
We run the experiments in batches.
For each pairs of models, we carry out $3-5$ human intelligence tests (HITs) on AMT.
Each HIT contains $70-90$ examples (dialogue context and two model responses) and is evaluated by $3-4$ unique humans.
In total we collect $5363$ preferences in $69$ HITs.

The following are screenshots from one actual Amazon Mechanical Turk (AMT) experiment. These screenshots show the introduction (debriefing) of the experiment, an example dialogue and one dialogue context with two candidate responses, which human evaluators were asked to choose between. The experiment was carried out using psiturk, which can be downloaded from \url{www.psiturk.org}. The source code will be released upon publication.

\begin{figure}[ht]
  \centering
  \includegraphics[width=1.0\linewidth]{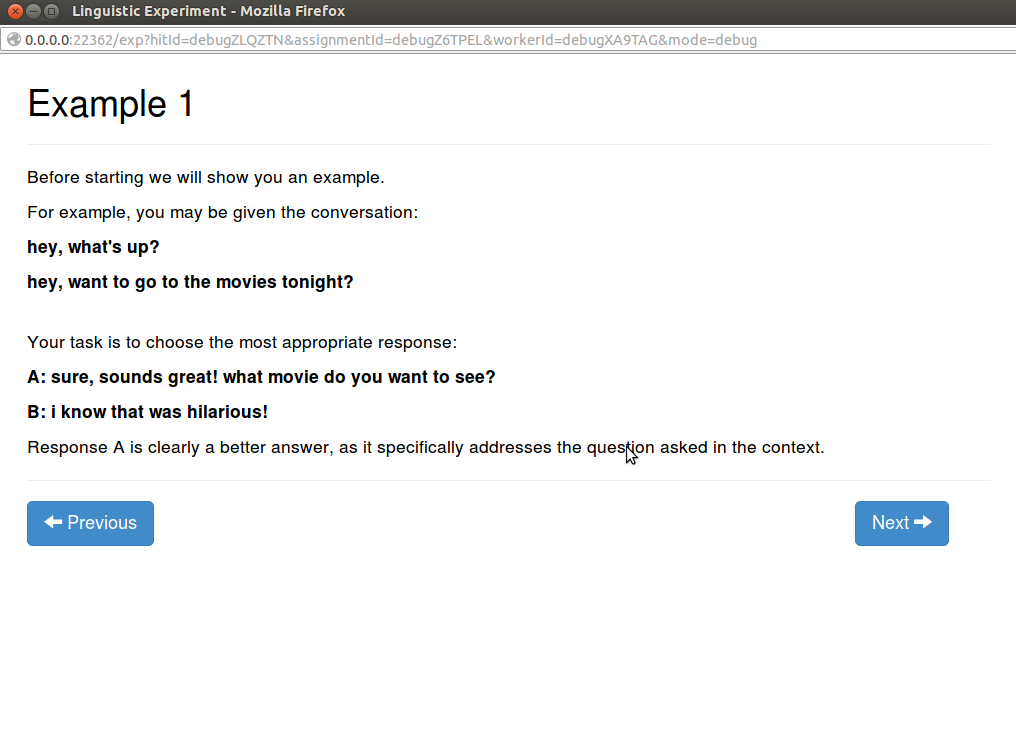}
  \caption{Screenshot of the introduction (debriefing) of the experiment.}
\end{figure}

\begin{figure}[ht]
  \centering
  \includegraphics[width=1.0\linewidth]{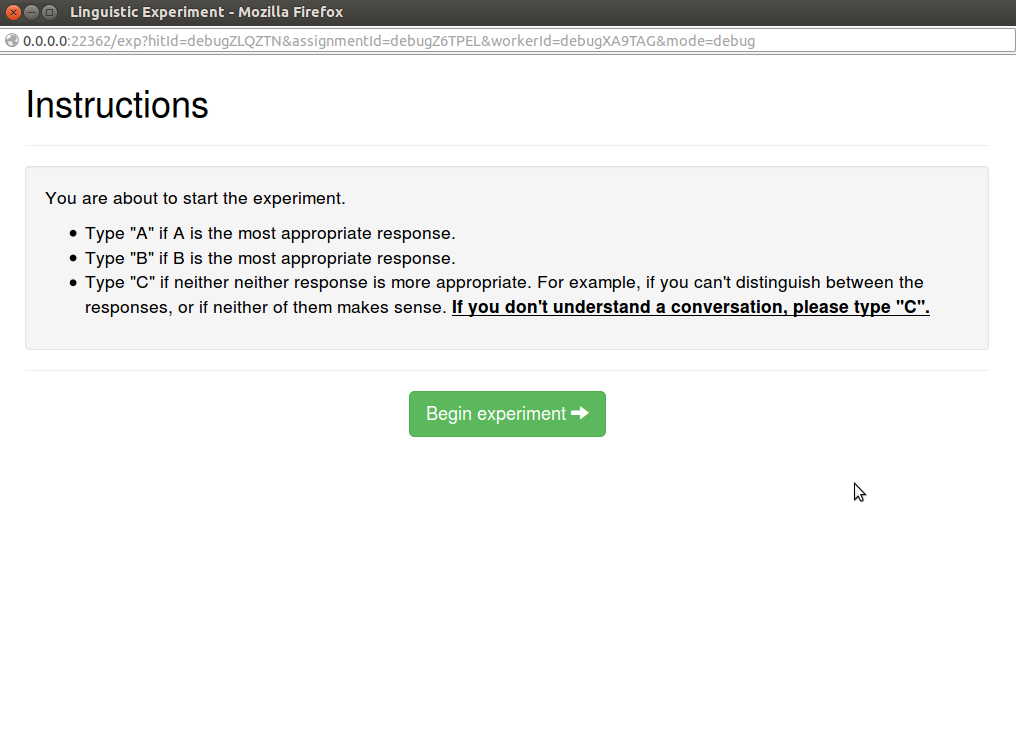}
  \caption{Screenshot of the introductory dialogue example.}
\end{figure}

\begin{figure}[ht]
  \centering
  \includegraphics[width=1.0\linewidth]{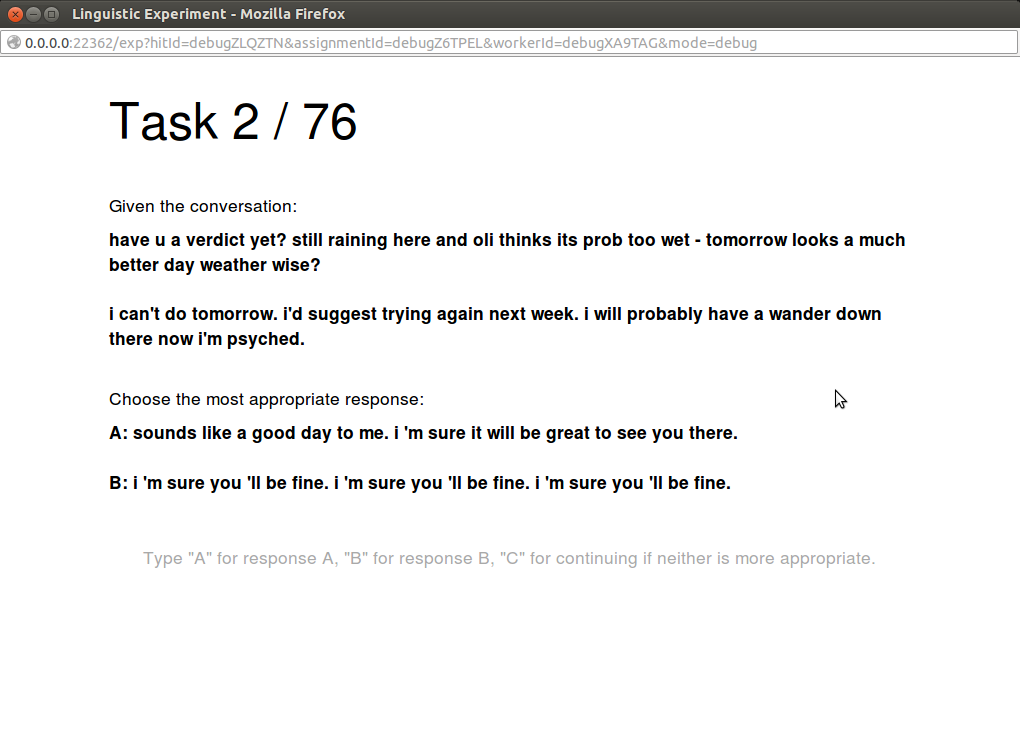}
  \caption{Screenshot of one dialogue context with two candidate responses, which human evaluators were asked to choose between.}
\end{figure}

\end{document}